\documentclass{article}

\PassOptionsToPackage{numbers, compress}{natbib}

 \usepackage[preprint]{neurips_2026}

\usepackage[utf8]{inputenc} 
\usepackage[T1]{fontenc}    
\usepackage{hyperref}       
\usepackage{url}            
\usepackage{booktabs}       
\usepackage{amsfonts}       
\usepackage{nicefrac}       
\usepackage{microtype}      
\usepackage{xcolor}         
\usepackage{multirow}
\usepackage{float}
\usepackage{wrapfig}
\usepackage{graphicx}
\usepackage{subcaption}
\usepackage{enumitem}

\usepackage{mathtools} 
\usepackage{amsmath}
\usepackage{amssymb}
\usepackage{amsthm}

\usepackage{algorithm}
\usepackage{algpseudocode}
\algdef{SE}[PAR]{ParFor}{EndParFor}
{\textbf{parallel for }}{\textbf{end parallel for}}

\title{DCD-PFN: A Decoupling-Aware Foundation Model for Causal Discovery}

\author{%
  \textbf{Zhengkang~Guan}, \textbf{Yikang~Chen}, \textbf{Yi~He}, \textbf{Yunze~Tong}, \textbf{Zijing~Hu},\\\textbf{Haoyuan~Qian}, \textbf{Fei~Wu}, \textbf{Kun~Kuang}\thanks{Corresponding author}\\
  College of Computer Science and Technology\\
  Zhejiang University\\
  \texttt{zhengkang.guan@zju.edu.cn, kunkuang@zju.edu.cn} \\
}

\begin{document}

\maketitle

\begin{abstract}
  Causal discovery is critical for understanding complex data-generating mechanisms, yet traditional algorithms often struggle with highly non-linear and noisy systems, or suffer from severe computational bottlenecks.
  Recent tabular foundation models based on Prior-Data Fitted Networks (PFNs) have demonstrated remarkable zero-shot inference capabilities, but their potential for explicit structural causal discovery remains underexplored.
  To bridge this gap, we propose DCD-PFN, a decoupling-aware foundation model for causal discovery.
  Instead of directly amortizing global graph reconstruction, DCD-PFN focuses on local causal discovery through a decoupling-based paradigm.
  Through pre-training on diverse synthetic Structural Causal Models (SCMs), the model learns sample-wise decoupling weights that enable Markov boundary (MB) identification.
  Furthermore, by leveraging parallelized local discovery, DCD-PFN efficiently reconstructs global causal graphs while remaining grounded in the theoretical foundations of decoupling-based causal discovery.
  Experiments demonstrate that our foundation model achieves robust zero-shot generalization.
\end{abstract}

\section{Introduction}

Causal discovery serves as the cornerstone for understanding the generative mechanisms underlying complex data and for enabling reliable decision-making.
In recent decades, traditional causal discovery algorithms have evolved into diverse paradigms, including constraint-based, score-based, and functional-based methods.
While these approaches are effective when their specific structural or distributional assumptions are satisfied, real-world data are typically characterized by highly complex relationships and noisy mechanisms that easily violate these assumptions.
Although constraint-based methods equipped with modern non-parametric conditional independence tests (CITs) offer a theoretically sound solution with minimal assumptions, they often suffer from prohibitive computational bottlenecks.
Consequently, developing a causal discovery framework that delivers both broad applicability and efficient inference remains a pressing challenge.

Recently, pre-trained tabular foundation models, represented by Prior-Data Fitted Networks (PFNs), have provided a novel perspective to address these challenges. By pre-training on massive amounts of synthetic data, these models (e.g., TabPFN~\cite{hollmann_accurate_2025}, LimiX~\cite{DBLP:journals/corr/abs-2509-03505}) have demonstrated remarkable in-context learning capabilities and zero-shot generalization performance on tabular prediction tasks, requiring only amortized forward-pass inference. However, existing PFN models are primarily designed as powerful posterior predictive distribution fitters. While this paradigm offers remarkable computational efficiency, its potential for explicit structural causal discovery with broad applicability and theoretical grounding remains largely underexplored.

To bridge this gap, we introduce DCD-PFN, a foundation model for causal discovery built upon a decoupling-based paradigm. Rather than amortizing global graph reconstruction, DCD-PFN learns sample-wise decoupling weights and then directly identifies Markov boundary (MB). By pre-training on diverse synthetic Structural Causal Models (SCMs), the model achieves powerful in-context learning capabilities for MB identification. Furthermore, through parallelized local discovery, DCD-PFN efficiently reconstructs the global causal graph while remaining strictly consistent with the theoretical framework of decoupling-based causal discovery~\citep{DCD2026}.

The main contributions of this paper are summarized as follows:
\begin{itemize}[topsep=0pt, left=5pt]
    \item We propose DCD-PFN, a foundation model specifically designed for explicit structural causal discovery, enabling out-of-the-box causal graph reconstruction via amortized inference.
    \item By adopting a parallelized decoupling-based local-to-global paradigm, DCD-PFN is not only anchored in a theoretical framework without restrictive assumptions, but also achieves highly efficient causal discovery.
    \item Experiments on both synthetic and real-world datasets demonstrate that DCD-PFN exhibits strong robustness and zero-shot generalization.
\end{itemize}

\section{Related Work}

\subsection{Prior-Data Fitted Networks}

Prior-Data Fitted Networks (PFNs) redefine Bayesian posterior predictive distribution estimation as in-context learning, utilizing Transformers~\cite{kossen2021selfattention} trained on synthetic priors to approximate predictive distributions~\cite{muller2022transformers, Stat}. This paradigm has been successfully validated on tabular data, outperforming tree-based baselines by leveraging priors derived from SCMs~\cite{hollmann2023tabpfn,hollmann_accurate_2025}. Recent efforts have further scaled this approach: TabPFN v2.5~\cite{grinsztajn2025tabpfn25advancingstateart} enhances context capacity via distillation, LimiX~\cite{DBLP:journals/corr/abs-2509-03505} enables joint distribution modeling, and Mitra~\cite{zhang2025mitra} optimizes prior diversity.

Beyond architecture scaling, concurrent research scrutinizes the reliability of PFNs under non-stationary settings~\cite{Closer2025, 2026realistic, 2026robustness}, with methods like Drift-Resilient TabPFN~\cite{Drift2024} incorporating time-variant SCM priors to address distribution shifts. Critically, this paradigm has also demonstrated significant potential in causality. However, while works such as CausalFM~\cite{anonymous2026foundation} and Do-PFN~\cite{robertson2025dopfn} show that PFNs can effectively model intervention outcomes, they remain primarily focused on approximating interventional distributions through black-box fitting. The potential of PFNs to serve as zero-shot foundation models for explicit structural causal discovery remains largely underexplored.

\subsection{Classical Causal and Markov Boundary Discovery}

Causal discovery seeks to reconstruct directed acyclic graphs (DAGs) from observational data. Traditional global structure learning methods often rely on conditional independence constraints (e.g., PC~\cite{spirtes1991algorithm}) or functional asymmetries (e.g., LiNGAM~\cite{shimizu2006linear}). Alternatively, local discovery focuses on the MB, the minimal set rendering a target independent of all other variables~\cite{JMLR:v11:aliferis10a}. Classical algorithms, such as Grow-Shrink (GS)~\cite{NIPS1999_5d79099f} and Incremental Association (IA)~\cite{Tsamardinos2003TimeAS}, alongside modern variants (BAMB~\cite{DBLP:journals/tist/LingYWLDW19}, EEMB~\cite{DBLP:journals/isci/WangLYW20}, CCMB~\cite{DBLP:journals/tcyb/WuJYMC20}), typically iteratively conduct conditional independence (CI) tests to discover the MB. 

\subsection{Amortized Inference for Causal Discovery}

To bypass the sample inefficiency and high latency of iterative algorithms, recent works have shifted towards amortized causal discovery. Leveraging the Dual-Axis Attention Block (DAB)~\cite{kossen2021selfattention}, which attends across both features and datapoints, models like AVICI~\cite{NEURIPS2022_54f7125d} and BCNP~\cite{dhir2025a} treat causal discovery as a supervised meta-learning problem. By pre-training on synthetic datasets generated from diverse SCMs, these models map observational data directly to causal graphs via feed-forward inference. This amortized synergy has also been extended to causal induction~\cite{ke2023learning}, active learning~\cite{annadani2024amortized}, and conditional independence testing~\cite{duong2025amortized}.

However, existing global amortized models often struggle with precision when reconstructing the entire graph simultaneously.
In contrast, our proposed framework bridges the gap between precise local discovery and rapid amortized inference. By focusing the foundation model strictly on decoupled MB discovery, we maintain the robustness of local methods while achieving zero-shot inference speed, providing a highly scalable module for subsequent global graph reconstruction.

\section{Method}

\subsection{Problem Setup}

Consider $d$ random variables $\mathbf{X} = (X_1, \dots, X_d) \in \mathcal{X} \subset \mathbb{R}^d$ and an observational dataset $\mathcal{D}_n = \{\mathbf{x}^{(i)}\}_{i=1}^n$ consisting of $n$ independent and identically distributed (i.i.d.) samples drawn from an underlying joint distribution $\mathbb{P}$. We assume the data-generating distribution $\mathbb{P}$ is Markovian with respect to a DAG $\mathcal{G} = (\mathcal{V}, E)$, where $\mathcal{V} = \{X_1, \dots, X_d\}$ serves as the set of nodes and $E$ denotes the set of directed edges. This implies that the joint distribution factorizes as:
\begin{equation}
    \mathbb{P}(X_1, \dots, X_d) = \prod_{j=1}^d \mathbb{P}(X_j \mid \text{PA}^{\mathcal{G}}(X_j)),
\end{equation}
where $\text{PA}^{\mathcal{G}}(X_j)$ denotes the set of parents of node $X_j$ in $\mathcal{G}$. Within the Bayesian framework, we treat both the DAG $\mathcal{G}$ and the model parameters $\theta$ as random variables. The joint prior distribution is specified as $\Pi(\mathcal{G}, \theta) = \Pi(\theta \mid \mathcal{G})\Pi(\mathcal{G})$, and the data-generating process is modeled as $\mathbf{x}^{(i)} \mid \mathcal{G}, \theta \overset{\text{i.i.d.}}{\sim} \mathbb{P}(\cdot \mid \mathcal{G}, \theta)$.

Amortized inference approaches frame causal structure learning as a supervised learning task. Instead of searching over the superexponential space of potential graphs, these methods parameterize a neural network $p_\phi(\mathcal{G} \mid \mathcal{D}_n)$ to directly approximate the intractable posterior $p(\mathcal{G} \mid \mathcal{D}_n)$. The optimal parameters $\phi^*$ are obtained by minimizing the expected negative log-likelihood (NLL) over the data-generating distribution:
\begin{equation}
\phi^* = \arg \min_{\phi} \mathbb{E}_{(\mathcal{G}, \theta) \sim \Pi} \mathbb{E}_{\mathcal{D}_n \sim \mathbb{P}(\cdot \mid \mathcal{G}, \theta)} \left[-\log p_\phi(\mathcal{G} \mid \mathcal{D}_n)\right].
\end{equation}
While directly learning the full graph $\mathcal{G}$ provides a complete picture of the system, it is often challenging and lacks identifiability guarantees. To address this, we first focus on the amortized learning of the MB of $\mathcal{G}$, denoted by $\text{MB}^{\mathcal{G}}(X_t)$, where $X_t$ denotes a specific target variable of interest, $t \in \{1, \dots, d\}$:
\begin{equation}
\eta^* = \arg \min_{\eta} \mathbb{E}_{(\mathcal{G}, \theta) \sim \Pi} \mathbb{E}_{\mathcal{D}_n \sim \mathbb{P}(\cdot \mid \mathcal{G}, \theta)} [-\log p_\eta(\text{MB}^{\mathcal{G}}(X_t) \mid \mathcal{D}_n)],
\end{equation}
where the joint prior $\Pi(\mathcal{G}, \theta) = \Pi(\theta \mid \mathcal{G})\Pi(\mathcal{G})$ is implicitly defined by the synthetic data generation mechanism employed during pre-training. By exposing the foundation model $p_\eta(\text{MB}^{\mathcal{G}}(X_t) \mid \mathcal{D}_n)$ to synthetic datasets generated by diverse SCMs, it learns to directly approximate the posterior distribution of the MB. Ultimately, by leveraging its end-to-end MB identification capabilities, we can iteratively recover the global graph structure.

\begin{figure}[t]
    \centering
    \includegraphics[width=1\linewidth]{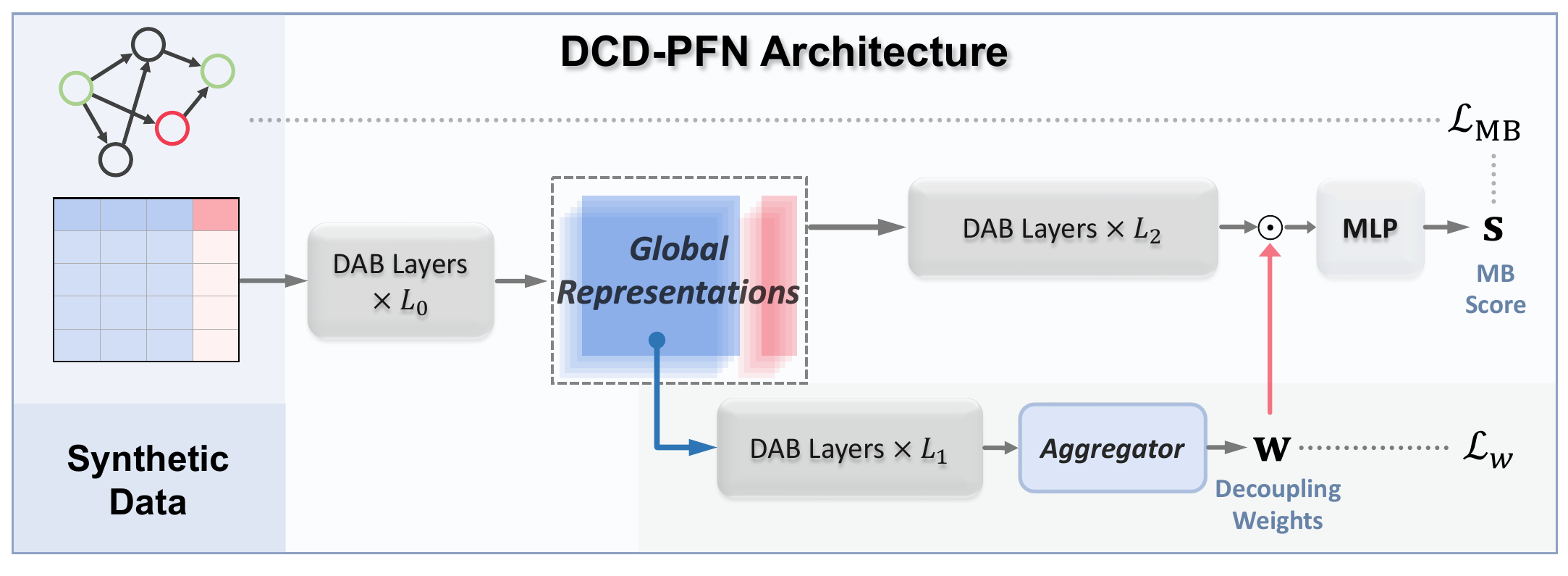}
    \caption{Architecture of DCD-PFN. Input synthetic data is encoded into global representations via $L_0$ Dual-Axis Attention Block (DAB) layers. The architecture then splits into two joint branches: (1) a decoupling branch that learns sample-wise weights $\mathbf{w}$ via $L_1$ DAB layers and an aggregator under the guidance of $\mathcal{L}_w$, and (2) an MB learning branch that refines representations via $L_2$ DAB layers and utilizes an MLP to output the final Markov Boundary scores $\mathbf{S}$ supervised by $\mathcal{L}_{MB}$.}\label{fig:main}
\end{figure}

\subsection{Model Architecture}

The foundational component of DCD-PFN is the Dual-Axis Attention Block (DAB)~\cite{kossen2021selfattention}, an architecture proven effective in capturing complex data dependencies by alternating between Feature Attention and Sample Attention. By explicitly modeling both the interactions between different features within a single observation and the relationships across different observations in the entire dataset, the DAB allows DCD-PFN to learn highly expressive representations.

The inputs $\mathbf{D} \in \mathbb{R}^{n \times d}$ are first mapped into an $e$-dimensional embedding space via linear projection, supplemented with unique feature ID embeddings to maintain column permutation invariance. An encoder consisting of $L_0$ DAB layers maps these inputs into a high-dimensional global representation space, yielding $\mathbf{H} \in \mathbb{R}^{n \times d \times e}$. Within this representation, the features excluding the target variable $X_t$ are denoted as $\mathbf{H}_{-t} \in \mathbb{R}^{n \times (d-1) \times e}$.

\textbf{Feature-wise Decoupling:} $\mathbf{H}_{-t}$ is utilized to learn sample decoupling weights $w_i, i \in \{1, \dots, n\}$ to decouple variables other than $X_t$. Specifically, $\mathbf{H}_{-t}$ is first encoded through $L_1$ DAB layers to obtain $\mathbf{H}^{(1)}_{-t} \in \mathbb{R}^{n \times (d-1) \times e}$, and is then processed by a cross-attention-based aggregator to compress the feature dimension. Using a learnable query token $\mathbf{q}$, the compressed representation $\mathbf{z}_i$ for each sample is computed as follows:
\begin{equation}
    \mathbf{z}_i = \text{LayerNorm}(\text{MHA}(\mathbf{q}, \mathbf{H}^{(1)}_{i, -t, \cdot}, \mathbf{H}^{(1)}_{i, -t, \cdot})),
\end{equation}
where $\text{MHA}$ denotes Multi-Head Attention. The representation $\mathbf{z}_i$ is subsequently mapped through a linear layer to output the final sample weight $w_i$.
    
\textbf{MB Learning:} The complete representation $\mathbf{H}$, combined with the learned decoupling weights $w_i, i \in \{1, \dots, n\}$, is employed to identify the $\text{MB}^{\mathcal{G}}(X_t)$. First, $\mathbf{H}$ passes through $L_2$ DAB layers to obtain a refined representation $\mathbf{H}^{(2)}$. Utilizing the learned decoupling weights, we then compute the weighted similarity between the target embeddings $\mathbf{H}^{(2)}_{t}$ and other variable embeddings $\mathbf{H}^{(2)}_{-t}$:
\begin{equation}
    \mathbf{S} = \sum_{i=1}^{n} w_i \cdot (\mathbf{H}^{(2)}_{i,-t} \odot \mathbf{H}^{(2)}_{i,t}), \quad \mathbf{S}_{t} = \sum_{i=1}^{n} w_i \cdot (\mathbf{H}^{(2)}_{i,t} \odot \mathbf{H}^{(2)}_{i,t}), \quad \mathbf{S}_{-t} = \sum_{i=1}^{n} w_i \cdot (\mathbf{H}^{(2)}_{i,-t} \odot \mathbf{H}^{(2)}_{i,-t}).
\end{equation}
Here, $\mathbf{S}$ captures the weighted cross-correlation in the embedding space, while $\mathbf{S}_{t}$ and $\mathbf{S}_{-t}$ serve as normalization metrics that account for target and other variables intensities. To ensure robustness across different magnitudes, these metrics undergo a signed logarithmic transformation:
\begin{equation}
    \mathbf{S}_{comb} = \text{Concat}(\text{sign}(\mathbf{S})\log(1+|\mathbf{S}|), \log(1+\mathbf{S}_{t}), \log(1+\mathbf{S}_{-t})).
\end{equation}
The concatenated vector is processed by an MLP with a Sigmoid activation function to yield the final probability score $\mathbf{s} \in [0, 1]^d$ representing the score of each variable belonging to the MB.

\subsection{Pre-training}

\textbf{Synthetic Data Generation:} The generalization capability of a causal foundation model relies heavily on the diversity of its pre-training prior. We generate synthetic priors using a vast scale of SCMs. For each dataset, the underlying DAG is sampled using the Erdős-Rényi model. The structural equations governing node relationships are stochastically selected from a diverse set of functional components, including random neural networks with varied activation functions, decision tree structures, and discretization steps. Finally, noise from multiple distributions with randomized variances is superimposed.

\begin{algorithm}[t]
\caption{Parallel DCD-Construction}
\label{alg:pdcd}
\begin{algorithmic}[1]
\Require Set of variables $\mathcal{V} = \{X_1, \dots, X_d\}$, MB identifier $\text{MB}(\cdot \mid \cdot)$
\Ensure Completed Partially Directed Acyclic Graph (CPDAG) $\hat{\mathcal{G}} = (\mathcal{V}, \hat{E})$
\State Initialize graph $\hat{\mathcal{G}} = (\mathcal{V}, \hat{E})$ with empty edge set $\hat{E} \leftarrow \emptyset$
\State Initialize v-structure set $\mathcal{K} \leftarrow \emptyset$
\ParFor{$X_t \in \mathcal{V}$} \Comment{parallel execution via native batch size $>1$ support}
    \State Obtain initial MB: $\mathcal{M}_t \leftarrow \text{MB}(X_t \mid \mathcal{V})$
    \State Initialize neighbors set $\mathcal{N}_t \leftarrow \mathcal{M}_t$
    \For{$k = 1$ to $|\mathcal{M}_t| - 1$}
        \For{each subset $\mathcal{Z} \subset \mathcal{M}_t$ with $|\mathcal{Z}| = k$}
            \State $\mathcal{M}_{\setminus \mathcal{Z}} \leftarrow \text{MB}(X_t \mid (\mathcal{M}_t \cup \{X_t\}) \setminus \mathcal{Z})$
            \State $\mathcal{D}_{\text{drop}} \leftarrow (\mathcal{M}_t \setminus \mathcal{Z}) \setminus \mathcal{M}_{\setminus \mathcal{Z}}$
            \If{$\mathcal{D}_{\text{drop}} \neq \emptyset$} 
                \For{each $Y \in \mathcal{D}_{\text{drop}}$} 
                    \For{each $Z \in \mathcal{Z}$}
                        \State $\mathcal{K} \leftarrow \mathcal{K} \cup \{(X_t, Z, Y)\}$ 
                    \EndFor
                    \State $\mathcal{N}_t \leftarrow \mathcal{N}_t \setminus \{Y\}$ 
                \EndFor
            \EndIf
        \EndFor
    \EndFor
    \For{each $W \in \mathcal{N}_t$}
        \State Add undirected edge to $\hat{E}$: $X_t - W$
    \EndFor
\EndParFor
\For{each $\{(X, Z, Y)\} \in \mathcal{K}$}
    \If{$(X - Z \in \hat{E}) \land (Y - Z \in \hat{E})$}
        \State Orient edges in $\hat{E}$ as: $X \rightarrow Z$ and $Y \rightarrow Z$
    \EndIf
\EndFor
\State Apply Meek rules to $\hat{\mathcal{G}}$ to exhaustively orient remaining undirected edges
\Return $\hat{\mathcal{G}}$
\end{algorithmic}
\end{algorithm}

The end-to-end model is optimized via a joint loss function:

\textbf{Decoupling Loss ($\mathcal{L}_{w}$):} We employ the Pairwise Weighted Hilbert-Schmidt Independence Criterion (HSIC) loss \citep{DCD2026}. Given the sample decoupling weights vector $\mathbf{w} = (w_1, \dots, w_n) \in \mathbb{R}^n$, we compute the initial kernel matrix $\mathbf{K}^{(j)} \in \mathbb{R}^{n \times n}$ for any non-target feature $X_j$ (where $j \not = t$) using a Radial Basis Function (RBF). Subsequently, a weighted centering operation is applied to the kernel matrix using normalized weights $\tilde{\mathbf{w}} = \mathbf{w} / \sum_{i=1}^n w_i$:
\begin{equation}
    \tilde{\mathbf{K}}^{(j)} = (\mathbf{I} - \mathbf{1}\tilde{\mathbf{w}}^T) \mathbf{K}^{(j)} (\mathbf{I} - \tilde{\mathbf{w}}\mathbf{1}^T)
\end{equation}
where $\mathbf{I}$ is the identity matrix and $\mathbf{1}$ is a vector of ones. To inject the absolute intensity of sample weights into the independence measure, we apply trace weighting to the centered kernel matrix:
\begin{equation}
    \bar{\mathbf{K}}^{(j)} = \mathbf{W}^{1/2} \tilde{\mathbf{K}}^{(j)} \mathbf{W}^{1/2}
\end{equation}
where $\mathbf{W} = \text{diag}(\mathbf{w})$ is the diagonal matrix constructed from the sample weights.

To eliminate the inherent scale differences between the kernel matrices of different feature dimensions, we compute the normalized HSIC (i.e., Centered Kernel Alignment). The final decoupling loss is defined as the sum of the normalized HSIC scores across all feature pairs:
\begin{equation}
    \mathcal{L}_{w} = \sum_{\substack{j < k \\ j, k \neq t}} \frac{\text{Tr}(\bar{\mathbf{K}}^{(j)} \bar{\mathbf{K}}^{(k)})}{\sqrt{\text{Tr}(\bar{\mathbf{K}}^{(j)} \bar{\mathbf{K}}^{(j)}) \text{Tr}(\bar{\mathbf{K}}^{(k)} \bar{\mathbf{K}}^{(k)})}}
\end{equation}

\textbf{MB Loss ($\mathcal{L}_{\text{MB}}$):} A standard Binary Cross-Entropy (BCE) loss is utilized to align the predicted MB probability $s_j$ with the ground-truth label $g_j$ derived from the synthetic SCMs:
\begin{equation}
    \mathcal{L}_{\text{MB}} = -\frac{1}{d-1} \sum_{\substack{j=1 \\ j \neq t}}^{d} [g_j \log(s_j) + (1-g_j) \log(1-s_j)]
\end{equation}

\subsection{Global Graph Reconstruction} 

While our foundation model acts as a highly efficient engine for local MB discovery, constructing a complete global causal graph requires synthesizing these local structures. The theoretical completeness and identifiability of reconstructing a global causal graph from local MBs have been established in concurrent theoretical framework \citep{DCD2026}.
Under standard causal discovery assumptions, we employ the Decoupled Causal Discovery (DCD) \citep{DCD2026} procedure detailed in Algorithm \ref{alg:pdcd} to reconstruct the global structure from these local components. By systematically marginalizing out candidate subsets within the local MBs and observing the dropping out of spouse nodes, the algorithm identifies unshielded colliders (v-structures) and true structural neighbors (the skeleton). Finally, applying Meek's rules exhaustively orients the remaining undirected edges, ultimately yielding a Completed Partially Directed Acyclic Graph (CPDAG) $\hat{\mathcal{G}}$ that represents the Markov equivalence class of the underlying data-generating mechanism.

\section{Experiments}

\subsection{Experimental Setup}

To comprehensively evaluate the performance of our proposed method, we benchmark it against a diverse and representative set of causal discovery algorithms encompassing several established paradigms:

\begin{itemize}[topsep=0pt, left=5pt]
    \item Constraint-based methods: The PC algorithm~\citep{PC}. To maximize its performance, the algorithm is implemented with 
    the Kernel-based CIT (KCIT)~\citep{KCIT} and its ensemble variants~\citep{ECIT}.
    \item Score-based methods: Greedy Equivalence Search (GES)~\citep{GES}.
    \item Functional-based methods: ICALiNGAM~\citep{shimizu2006linear} and DirectLiNGAM~\citep{shimizu2011directlingam}.
    \item Continuous Optimization methods: NOTEARS~\citep{NOTEARS}, GraNDAG~\citep{GraNDAG}, and GOLEM~\citep{GOLEM}.
    \item Amortized Inference: AVICI~\citep{NEURIPS2022_54f7125d}, serving as an amortized transformer-based baseline.
\end{itemize}

Since the evaluated methods output varying causal structures, either DAGs or CPDAGs, we uniformly convert all estimated outputs into their corresponding CPDAGs for evaluation. This standardization ensures a fair and consistent comparison across all algorithmic paradigms.

We measure the structural discovery performance using standard evaluation metrics: Structural Hamming Distance adapted for CPDAGs ($\text{SHD}_c$), Precision, Recall, and F1-score.

\begin{table}[t]
\centering
\caption{Performance evaluation across different dimensions ($d \in \{10, 15, 20\}$)}
\label{tab:SF1}
\small
\begin{tabular}{lcccc}
\toprule
Method & SHD$_c \downarrow$ & Precision $\uparrow$ & Recall $\uparrow$ & F1 $\uparrow$ \\
\midrule
\multicolumn{5}{c}{\textbf{$d=10$}} \\
\midrule
DCD-PFN & \textbf{6.560 $\pm$ 1.121} & \textbf{0.949 $\pm$ 0.129} & \textbf{0.289 $\pm$ 0.116} & \textbf{0.431 $\pm$ 0.140} \\
AVICI & 7.440 $\pm$ 1.325 & 0.695 $\pm$ 0.459 & 0.173 $\pm$ 0.147 & 0.264 $\pm$ 0.208 \\
PC (KCIT) & 8.680 $\pm$ 1.492 & 0.152 $\pm$ 0.159 & 0.120 $\pm$ 0.128 & 0.133 $\pm$ 0.140 \\
PC (E-KCIT) & 7.960 $\pm$ 1.338 & 0.212 $\pm$ 0.199 & 0.142 $\pm$ 0.130 & 0.167 $\pm$ 0.151 \\
GES & 8.640 $\pm$ 2.196 & 0.230 $\pm$ 0.271 & 0.173 $\pm$ 0.198 & 0.196 $\pm$ 0.227 \\
ICALiNGAM & 7.560 $\pm$ 1.158 & 0.800 $\pm$ 0.408 & 0.160 $\pm$ 0.129 & 0.257 $\pm$ 0.180 \\
DirectLiNGAM & 7.560 $\pm$ 1.193 & 0.760 $\pm$ 0.436 & 0.160 $\pm$ 0.133 & 0.255 $\pm$ 0.188 \\
Notears & 8.240 $\pm$ 0.779 & 0.600 $\pm$ 0.500 & 0.084 $\pm$ 0.087 & 0.145 $\pm$ 0.140 \\
GraNDAG & 13.320 $\pm$ 4.905 & 0.323 $\pm$ 0.372 & 0.151 $\pm$ 0.106 & 0.174 $\pm$ 0.133 \\
GOLEM & 7.720 $\pm$ 0.980 & 0.800 $\pm$ 0.408 & 0.142 $\pm$ 0.109 & 0.234 $\pm$ 0.161 \\
\midrule
\multicolumn{5}{c}{\textbf{$d=15$}} \\
\midrule
DCD-PFN & \textbf{11.000 $\pm$ 1.581} & 0.883 $\pm$ 0.166 & \textbf{0.260 $\pm$ 0.092} & \textbf{0.391 $\pm$ 0.117} \\
AVICI & 12.560 $\pm$ 4.234 & 0.691 $\pm$ 0.397 & 0.186 $\pm$ 0.111 & 0.281 $\pm$ 0.167 \\
PC (KCIT) & 14.400 $\pm$ 1.633 & 0.106 $\pm$ 0.100 & 0.094 $\pm$ 0.089 & 0.100 $\pm$ 0.094 \\
PC (E-KCIT) & 12.880 $\pm$ 1.424 & 0.169 $\pm$ 0.139 & 0.114 $\pm$ 0.092 & 0.135 $\pm$ 0.110 \\
GES & 13.708 $\pm$ 2.476 & 0.205 $\pm$ 0.170 & 0.161 $\pm$ 0.130 & 0.178 $\pm$ 0.145 \\
ICALiNGAM & 11.920 $\pm$ 1.038 & 0.870 $\pm$ 0.302 & 0.151 $\pm$ 0.075 & 0.253 $\pm$ 0.114 \\
DirectLiNGAM & 11.360 $\pm$ 1.150 & 0.950 $\pm$ 0.204 & 0.191 $\pm$ 0.082 & 0.313 $\pm$ 0.118 \\
Notears & 12.640 $\pm$ 1.114 & 0.787 $\pm$ 0.407 & 0.100 $\pm$ 0.080 & 0.172 $\pm$ 0.127 \\
GraNDAG & 19.680 $\pm$ 6.866 & 0.386 $\pm$ 0.349 & 0.166 $\pm$ 0.087 & 0.187 $\pm$ 0.096 \\
GOLEM & 11.680 $\pm$ 1.030 & \textbf{0.990 $\pm$ 0.050} & 0.169 $\pm$ 0.074 & 0.281 $\pm$ 0.104 \\
\midrule
\multicolumn{5}{c}{\textbf{$d=20$}} \\
\midrule
DCD-PFN & \textbf{14.520 $\pm$ 2.740} & 0.826 $\pm$ 0.184 & \textbf{0.301 $\pm$ 0.101} & \textbf{0.435 $\pm$ 0.162} \\
AVICI & 21.760 $\pm$ 11.061 & 0.483 $\pm$ 0.395 & 0.173 $\pm$ 0.152 & 0.231 $\pm$ 0.189 \\
PC (KCIT) & 20.280 $\pm$ 2.923 & 0.125 $\pm$ 0.091 & 0.101 $\pm$ 0.076 & 0.111 $\pm$ 0.082 \\
PC (E-KCIT) & 18.040 $\pm$ 1.837 & 0.177 $\pm$ 0.121 & 0.109 $\pm$ 0.066 & 0.133 $\pm$ 0.082 \\
GES & 19.840 $\pm$ 3.118 & 0.170 $\pm$ 0.148 & 0.149 $\pm$ 0.131 & 0.159 $\pm$ 0.138 \\
ICALiNGAM & 16.200 $\pm$ 1.323 & 0.930 $\pm$ 0.221 & 0.149 $\pm$ 0.071 & 0.251 $\pm$ 0.108 \\
DirectLiNGAM & 15.640 $\pm$ 1.524 & 0.982 $\pm$ 0.070 & 0.183 $\pm$ 0.083 & 0.300 $\pm$ 0.113 \\
Notears & 17.120 $\pm$ 1.013 & 0.920 $\pm$ 0.277 & 0.099 $\pm$ 0.053 & 0.176 $\pm$ 0.089 \\
GraNDAG & 31.000 $\pm$ 19.636 & 0.290 $\pm$ 0.304 & 0.099 $\pm$ 0.046 & 0.123 $\pm$ 0.072 \\
GOLEM & 15.800 $\pm$ 1.414 & \textbf{0.983 $\pm$ 0.063} & 0.175 $\pm$ 0.080 & 0.288 $\pm$ 0.109 \\
\bottomrule
\end{tabular}
\end{table}

\subsection{Synthetic Datasets}

Following the data generation protocol outlined in \citet{DCD2026}, we construct synthetic datasets using continuous Structural Equation Models (SEMs) based on randomly generated DAGs.

Specifically, the underlying causal structures are simulated as Scale-Free (SF) networks using the Barabási-Albert model. We set the number of nodes $d \in \{10, 15, 20\}$ and attach each new node to one 
existing nodes. Random node permutation and topological sorting are applied to ensure acyclicity. The continuous data is then generated sequentially via the following additive noise model:
\begin{equation*}
    x_i = f(W_i^T x_{pa(i)} + b_i) + \epsilon_i
\end{equation*}
where $x_{pa(i)}$ denotes the feature vector of node $i$'s parents. Consistent with \citet{DCD2026}, the weight matrix $W_i$ is initialized using the Glorot uniform scheme, and the bias term $b_i$ is sampled from a standard normal distribution. To reflect non-linear causal mechanisms, the transformation function $f(\cdot)$ is uniformly sampled from the set $\{z, z^2, z^3, \tanh(z), \exp(z), \log(|z|)\}$. The intermediate output of each node is strictly standardized to zero mean and unit variance to prevent numerical overflow and maintain a stable signal scale.

Following \citet{DCD2026}, we model the additive noise $\epsilon_i$ using four standard distributions: Gaussian, Laplace, Uniform and Student's $t$. However, while \citet{DCD2026} focuses on systematically scaling the noise magnitude to evaluate noise robustness of different causal discovery methods, our primary objective is to assess the framework's performance under a unified challenging setting. Therefore, instead of varying the noise scales, we uniformly sample the standard deviation of $\epsilon_i$ from the interval $[1.0, 5.0]$ for each node independently. Finally, for each simulated causal graph, we draw $n = 800$ observational samples. To ensure the statistical reliability of our results, all synthetic experiments are independently repeated 25 times with different random seeds.

Crucially, we emphasize that our evaluation environments represent a strict \textbf{zero-shot} generalization challenge relative to the pre-training space of DCD-PFN. Specifically, the synthetic test scenarios introduce severe distribution shifts across three distinct dimensions:
\begin{itemize}[topsep=0pt, left=5pt]
    \item \textbf{Topological}: Generalizing from homogeneous Erdős-Rényi (ER) graphs seen during pre-training to highly skewed Scale-Free (SF) networks.
    \item \textbf{Noise Intensity}: Transitioning from standard low-variance environments to a challenging moderate-to-high regime.
    \item \textbf{Mechanistic}: Extrapolating from implicit neural network priors to explicit, highly non-linear functional mechanisms (e.g., high-degree polynomials and exponential functions).
\end{itemize}

As shown in Table~\ref{tab:SF1}, DCD-PFN consistently achieves the lowest $\text{SHD}_c$ and highest F1-scores, demonstrating superior structural robustness compared to traditional approaches. While methods like GOLEM yield marginally higher precision in $d=15$ and $d=20$, they suffer from severely compromised recall ($\sim 0.17$). In contrast, DCD-PFN achieves a much better balance, maintaining competitive precision while yielding a substantially higher recall ($\sim 0.30$), which leads to its dominant F1-performance. Notably, compared to AVICI, which shares the same pre-training paradigm, DCD-PFN exhibits a substantial performance advantage, with F1-score nearly 90\% higher across all settings.

\subsection{Real-world Datasets}

Following~\citet{ECIT}, we evaluate DCD-PFN on the Sachs Flow-Cytometry dataset~\citep{FlowCytometry}, a widely used benchmark in causal discovery. This dataset is a standard benchmark in causal discovery, containing measurements of $d=11$ phosphorylated proteins and phospholipids derived from primary human CD4+ T cells. The consensus causal graph, meticulously validated by domain experts in the original study, serves as our ground truth for evaluation.

Tables~\ref{tab:sachs1} summarizes the structural identification performance of the Sachs dataset. DCD-PFN delivers the best overall performance, capturing the absolute minimum structural error ($\text{SHD}_c = 14$) and achieving a perfect Precision of 1.000. By contrast, its pre-trained counterpart, AVICI, shows limited performance on this real-world dataset. The results validate the robust zero-shot generalization capabilities of DCD-PFN on real-world data.


\begin{table}[t]
\centering
\caption{Performance comparison on the Sachs Flow-Cytometry Dataset.}
\label{tab:sachs1}
\begin{tabular}{lccccc}
\toprule
Method & SHD$_c \downarrow$ & Precision $\uparrow$ & Recall $\uparrow$ & F1 $\uparrow$ \\
\midrule
DCD-PFN        & \textbf{14} & \textbf{1.000} & \textbf{0.300} & \textbf{0.462} \\
AVICI         & 19 & 0.286 & 0.100 & 0.148 \\
PC (KCIT)     & 19 & 0.333 & 0.200 & 0.250 \\
PC (E-KCIT)   & 16 & 0.545 & \textbf{0.300} & 0.387 \\
GES           & 19 & 0.308 & 0.200 & 0.242 \\
ICALiNGAM     & 15 & 0.714 & 0.250 & 0.370 \\
DirectLiNGAM  & 15 & 0.833 & 0.250 & 0.385 \\
Notears       & 18 & 0.500 & 0.150 & 0.231 \\
GraNDAG       & 20 & 0.000 & 0.000 & 0.000 \\
GOLEM         & 15 & 0.667 & \textbf{0.300} & 0.414 \\
\bottomrule
\end{tabular}
\end{table}

\section{Conclusion}

In this paper, we introduced DCD-PFN, a novel foundation model tailored for explicit structural causal discovery. By leveraging the powerful in-context learning capabilities of Prior-Data Fitted Networks on tabular data, DCD-PFN achieves end-to-end identification of local causal structures through a precise decoupling mechanism. To circumvent the inherent challenges of direct amortized inference for global graph reconstruction, we adopted a local-to-global approach. Through parallel and accelerated local discovery, our framework can efficiently recover the global causal graph structure while preserving identifiability guarantees. By pre-training on a diverse range of SCMs, DCD-PFN implicitly internalizes complex decoupling rules. Experiments on complex synthetic and real-world tabular datasets demonstrate that our foundation model achieves robust zero-shot generalization and out-of-the-box structural reconstruction.


\bibliography{references}

\begin{thebibliography}{37}
\providecommand{\natexlab}[1]{#1}
\providecommand{\url}[1]{\texttt{#1}}
\expandafter\ifx\csname urlstyle\endcsname\relax
  \providecommand{\doi}[1]{doi: #1}\else
  \providecommand{\doi}{doi: \begingroup \urlstyle{rm}\Url}\fi

\bibitem[Aliferis et~al.(2010)Aliferis, Statnikov, Tsamardinos, Mani, and Koutsoukos]{JMLR:v11:aliferis10a}
Constantin~F. Aliferis, Alexander Statnikov, Ioannis Tsamardinos, Subramani Mani, and Xenofon~D. Koutsoukos.
\newblock Local causal and markov blanket induction for causal discovery and feature selection for classification part i: Algorithms and empirical evaluation.
\newblock \emph{Journal of Machine Learning Research}, 11\penalty0 (7):\penalty0 171--234, 2010.

\bibitem[Annadani et~al.(2024)Annadani, Tigas, Bauer, and Foster]{annadani2024amortized}
Yashas Annadani, Panagiotis Tigas, Stefan Bauer, and Adam Foster.
\newblock Amortized active causal induction with deep reinforcement learning.
\newblock In \emph{ICML 2024 Workshop on Structured Probabilistic Inference {\&} Generative Modeling}, 2024.

\bibitem[Cheng et~al.(2025)Cheng, Jia, Zhou, Li, and Guo]{2026realistic}
Zi-Jian Cheng, Zi-Yi Jia, Zhi Zhou, Yu-Feng Li, and Lan-Zhe Guo.
\newblock Realistic evaluation of tabpfn v2 in open environments, 2025.

\bibitem[Chickering(2002)]{GES}
David~Maxwell Chickering.
\newblock Optimal structure identification with greedy search.
\newblock \emph{Journal of machine learning research}, 3\penalty0 (Nov):\penalty0 507--554, 2002.

\bibitem[Dhir et~al.(2025)Dhir, Ashman, Requeima, and van~der Wilk]{dhir2025a}
Anish Dhir, Matthew Ashman, James Requeima, and Mark van~der Wilk.
\newblock A meta-learning approach to bayesian causal discovery.
\newblock In \emph{The Thirteenth International Conference on Learning Representations}, 2025.

\bibitem[Djilani et~al.(2025)Djilani, Simonetto, Tit, Tambon, Récamier, Ghamizi, Cordy, and Papadakis]{2026robustness}
Mohamed Djilani, Thibault Simonetto, Karim Tit, Florian Tambon, Paul Récamier, Salah Ghamizi, Maxime Cordy, and Mike Papadakis.
\newblock On the robustness of tabular foundation models: Test-time attacks and in-context defenses, 2025.

\bibitem[Duong et~al.(2025)Duong, Hoang, and Nguyen]{duong2025amortized}
Bao Duong, Nu~Hoang, and Thin Nguyen.
\newblock Amortized conditional independence testing.
\newblock In Xintao Wu, Myra Spiliopoulou, Can Wang, Vipin Kumar, Longbing Cao, Yanqiu Wu, Yu~Yao, and Zhangkai Wu, editors, \emph{Advances in Knowledge Discovery and Data Mining}, pages 410--423, Singapore, 2025. Springer Nature Singapore.
\newblock ISBN 978-981-96-8170-9.

\bibitem[Grinsztajn et~al.(2025)Grinsztajn, Flöge, Key, Birkel, Jund, Roof, Jäger, Safaric, Alessi, Hayler, Manium, Yu, Jablonski, Hoo, Garg, Robertson, Bühler, Moroshan, Purucker, Cornu, Wehrhahn, Bonetto, Schölkopf, Gambhir, Hollmann, and Hutter]{grinsztajn2025tabpfn25advancingstateart}
Léo Grinsztajn, Klemens Flöge, Oscar Key, Felix Birkel, Philipp Jund, Brendan Roof, Benjamin Jäger, Dominik Safaric, Simone Alessi, Adrian Hayler, Mihir Manium, Rosen Yu, Felix Jablonski, Shi~Bin Hoo, Anurag Garg, Jake Robertson, Magnus Bühler, Vladyslav Moroshan, Lennart Purucker, Clara Cornu, Lilly~Charlotte Wehrhahn, Alessandro Bonetto, Bernhard Schölkopf, Sauraj Gambhir, Noah Hollmann, and Frank Hutter.
\newblock Tabpfn-2.5: Advancing the state of the art in tabular foundation models, 2025.

\bibitem[Guan and Kuang(2026{\natexlab{a}})]{DCD2026}
Zhengkang Guan and Kun Kuang.
\newblock Decoupled causal discovery, 2026{\natexlab{a}}.
\newblock Manuscript in preparation.

\bibitem[Guan and Kuang(2026{\natexlab{b}})]{ECIT}
Zhengkang Guan and Kun Kuang.
\newblock Efficient ensemble conditional independence test framework for causal discovery.
\newblock In \emph{International Conference on Learning Representations}, 2026{\natexlab{b}}.

\bibitem[Helli et~al.(2024)Helli, Schnurr, Hollmann, M\"{u}ller, and Hutter]{Drift2024}
Kai Helli, David Schnurr, Noah Hollmann, Samuel M\"{u}ller, and Frank Hutter.
\newblock Drift-resilient tabpfn: In-context learning temporal distribution shifts on tabular data.
\newblock In A.~Globerson, L.~Mackey, D.~Belgrave, A.~Fan, U.~Paquet, J.~Tomczak, and C.~Zhang, editors, \emph{Advances in Neural Information Processing Systems}, volume~37, pages 98742--98781. Curran Associates, Inc., 2024.

\bibitem[Hollmann et~al.(2023)Hollmann, M{\"u}ller, Eggensperger, and Hutter]{hollmann2023tabpfn}
Noah Hollmann, Samuel M{\"u}ller, Katharina Eggensperger, and Frank Hutter.
\newblock Tab{PFN}: A transformer that solves small tabular classification problems in a second.
\newblock In \emph{The Eleventh International Conference on Learning Representations}, 2023.

\bibitem[Hollmann et~al.(2025)Hollmann, Müller, Purucker, Krishnakumar, Körfer, Hoo, Schirrmeister, and Hutter]{hollmann_accurate_2025}
Noah Hollmann, Samuel Müller, Lennart Purucker, Arjun Krishnakumar, Max Körfer, Shi~Bin Hoo, Robin~Tibor Schirrmeister, and Frank Hutter.
\newblock Accurate predictions on small data with a tabular foundation model.
\newblock \emph{Nature}, 637\penalty0 (8045):\penalty0 319--326, January 2025.
\newblock ISSN 1476-4687.

\bibitem[Ke et~al.(2023)Ke, Chiappa, Wang, Bornschein, Goyal, Rey, Weber, Botvinick, Mozer, and Rezende]{ke2023learning}
Nan~Rosemary Ke, Silvia Chiappa, Jane~X Wang, Jorg Bornschein, Anirudh Goyal, Melanie Rey, Theophane Weber, Matthew Botvinick, Michael~Curtis Mozer, and Danilo~Jimenez Rezende.
\newblock Learning to induce causal structure.
\newblock In \emph{International Conference on Learning Representations}, 2023.

\bibitem[Kossen et~al.(2021)Kossen, Band, Lyle, Gomez, Rainforth, and Gal]{kossen2021selfattention}
Jannik Kossen, Neil Band, Clare Lyle, Aidan Gomez, Tom Rainforth, and Yarin Gal.
\newblock Self-attention between datapoints: Going beyond individual input-output pairs in deep learning.
\newblock In A.~Beygelzimer, Y.~Dauphin, P.~Liang, and J.~Wortman Vaughan, editors, \emph{Advances in Neural Information Processing Systems}, 2021.

\bibitem[Lachapelle et~al.(2019)Lachapelle, Brouillard, Deleu, and Lacoste-Julien]{GraNDAG}
S{\'e}bastien Lachapelle, Philippe Brouillard, Tristan Deleu, and Simon Lacoste-Julien.
\newblock Gradient-based neural dag learning.
\newblock \emph{ArXiv}, abs/1906.02226, 2019.

\bibitem[Ling et~al.(2019)Ling, Yu, Wang, Liu, Ding, and Wu]{DBLP:journals/tist/LingYWLDW19}
Zhaolong Ling, Kui Yu, Hao Wang, Lin Liu, Wei Ding, and Xindong Wu.
\newblock Bamb: A balanced markov blanket discovery approach to feature selection.
\newblock \emph{ACM Trans. Intell. Syst. Technol.}, 10\penalty0 (5):\penalty0 52:1--52:25, 2019.

\bibitem[Lorch et~al.(2022)Lorch, Sussex, Rothfuss, Krause, and Sch\"{o}lkopf]{NEURIPS2022_54f7125d}
Lars Lorch, Scott Sussex, Jonas Rothfuss, Andreas Krause, and Bernhard Sch\"{o}lkopf.
\newblock Amortized inference for causal structure learning.
\newblock In S.~Koyejo, S.~Mohamed, A.~Agarwal, D.~Belgrave, K.~Cho, and A.~Oh, editors, \emph{Advances in Neural Information Processing Systems}, volume~35, pages 13104--13118. Curran Associates, Inc., 2022.

\bibitem[Ma et~al.(2026)Ma, Frauen, Javurek, and Feuerriegel]{anonymous2026foundation}
Yuchen Ma, Dennis Frauen, Emil Javurek, and Stefan Feuerriegel.
\newblock Foundation models for causal inference via prior-data fitted networks.
\newblock In \emph{The Fourteenth International Conference on Learning Representations}, 2026.

\bibitem[Margaritis and Thrun(1999)]{NIPS1999_5d79099f}
Dimitris Margaritis and Sebastian Thrun.
\newblock Bayesian network induction via local neighborhoods.
\newblock In S.~Solla, T.~Leen, and K.~M\"{u}ller, editors, \emph{Advances in Neural Information Processing Systems}, volume~12. MIT Press, 1999.

\bibitem[M{\"u}ller et~al.(2022)M{\"u}ller, Hollmann, Arango, Grabocka, and Hutter]{muller2022transformers}
Samuel M{\"u}ller, Noah Hollmann, Sebastian~Pineda Arango, Josif Grabocka, and Frank Hutter.
\newblock Transformers can do bayesian inference.
\newblock In \emph{International Conference on Learning Representations}, 2022.

\bibitem[Nagler(2023)]{Stat}
Thomas Nagler.
\newblock Statistical foundations of prior-data fitted networks.
\newblock In \emph{Proceedings of the 40th International Conference on Machine Learning}, ICML'23. JMLR.org, 2023.

\bibitem[Ng et~al.(2020)Ng, Ghassami, and Zhang]{GOLEM}
Ignavier Ng, AmirEmad Ghassami, and Kun Zhang.
\newblock On the role of sparsity and dag constraints for learning linear dags.
\newblock \emph{ArXiv}, abs/2006.10201, 2020.

\bibitem[Robertson et~al.(2025)Robertson, Reuter, Guo, Hollmann, Hutter, and Sch{\"o}lkopf]{robertson2025dopfn}
Jake Robertson, Arik Reuter, Siyuan Guo, Noah Hollmann, Frank Hutter, and Bernhard Sch{\"o}lkopf.
\newblock Do-{PFN}: In-context learning for causal effect estimation.
\newblock In \emph{The Thirty-ninth Annual Conference on Neural Information Processing Systems}, 2025.

\bibitem[Sachs et~al.(2005)Sachs, Perez, Pe'er, Lauffenburger, and Nolan]{FlowCytometry}
Karen Sachs, Omar Perez, Dana Pe'er, Douglas~A Lauffenburger, and Garry~P Nolan.
\newblock Causal protein-signaling networks derived from multiparameter single-cell data.
\newblock \emph{Science}, 308\penalty0 (5721):\penalty0 523--529, 2005.

\bibitem[Shimizu et~al.(2006)Shimizu, Hoyer, Hyv{\"a}rinen, Kerminen, and Jordan]{shimizu2006linear}
Shohei Shimizu, Patrik~O Hoyer, Aapo Hyv{\"a}rinen, Antti Kerminen, and Michael Jordan.
\newblock A linear non-gaussian acyclic model for causal discovery.
\newblock \emph{Journal of Machine Learning Research}, 7\penalty0 (10), 2006.

\bibitem[Shimizu et~al.(2011)Shimizu, Inazumi, Sogawa, Hyvarinen, Kawahara, Washio, Hoyer, Bollen, and Hoyer]{shimizu2011directlingam}
Shohei Shimizu, Takanori Inazumi, Yasuhiro Sogawa, Aapo Hyvarinen, Yoshinobu Kawahara, Takashi Washio, Patrik~O Hoyer, Kenneth Bollen, and Patrik Hoyer.
\newblock Directlingam: A direct method for learning a linear non-gaussian structural equation model.
\newblock \emph{Journal of Machine Learning Research-JMLR}, 12\penalty0 (Apr):\penalty0 1225--1248, 2011.

\bibitem[Spirtes and Glymour(1991)]{spirtes1991algorithm}
Peter Spirtes and Clark Glymour.
\newblock An algorithm for fast recovery of sparse causal graphs.
\newblock \emph{Social Science Computer Review}, 9\penalty0 (1):\penalty0 62--72, 1991.

\bibitem[Spirtes et~al.(2000)Spirtes, Glymour, and Scheines]{PC}
Peter Spirtes, Clark~N Glymour, and Richard Scheines.
\newblock \emph{Causation, prediction, and search}.
\newblock MIT press, 2000.

\bibitem[Tsamardinos et~al.(2003)Tsamardinos, Aliferis, and Statnikov]{Tsamardinos2003TimeAS}
I.~Tsamardinos, Constantin~F. Aliferis, and Alexander~R. Statnikov.
\newblock Time and sample efficient discovery of markov blankets and direct causal relations.
\newblock In \emph{Knowledge Discovery and Data Mining}, 2003.

\bibitem[Wang et~al.(2020)Wang, Ling, Yu, and Wu]{DBLP:journals/isci/WangLYW20}
Hao Wang, Zhaolong Ling, Kui Yu, and Xindong Wu.
\newblock Towards efficient and effective discovery of markov blankets for feature selection.
\newblock \emph{Inf. Sci.}, 509:\penalty0 227--242, 2020.

\bibitem[Wu et~al.(2020)Wu, Jiang, Yu, Miao, and Chen]{DBLP:journals/tcyb/WuJYMC20}
Xingyu Wu, Bingbing Jiang, Kui Yu, Chunyan Miao, and Huanhuan Chen.
\newblock Accurate markov boundary discovery for causal feature selection.
\newblock \emph{IEEE Trans. Cybern.}, 50\penalty0 (12):\penalty0 4983--4996, 2020.

\bibitem[Ye et~al.(2025)Ye, Liu, and Chao]{Closer2025}
Han-Jia Ye, Si-Yang Liu, and Wei-Lun Chao.
\newblock A closer look at tab{PFN} v2: Understanding its strengths and extending its capabilities.
\newblock In \emph{The Thirty-ninth Annual Conference on Neural Information Processing Systems}, 2025.

\bibitem[Zhang et~al.(2011)Zhang, Peters, Janzing, and Sch{\"{o}}lkopf]{KCIT}
Kun Zhang, Jonas Peters, Dominik Janzing, and Bernhard Sch{\"{o}}lkopf.
\newblock Kernel-based conditional independence test and application in causal discovery.
\newblock In F{\'{a}}bio~Gagliardi Cozman and Avi Pfeffer, editors, \emph{{UAI} 2011, Proceedings of the Twenty-Seventh Conference on Uncertainty in Artificial Intelligence, Barcelona, Spain, July 14-17, 2011}, pages 804--813. {AUAI} Press, 2011.

\bibitem[Zhang et~al.(2025{\natexlab{a}})Zhang, Ren, Yu, Yuan, Wang, Li, Wu, Mo, Mao, Hao, Dai, Xu, Li, Zhang, He, Wang, Zhang, Xu, Li, Gao, Zou, Liu, Liu, Xu, Cheng, Li, Zhou, Li, Fan, Lin, Han, Li, Lu, Xue, Jiang, Wang, Wang, and Cui]{DBLP:journals/corr/abs-2509-03505}
Xingxuan Zhang, Gang Ren, Han Yu, Hao Yuan, Hui Wang, Jiansheng Li, Jiayun Wu, Lang Mo, Li~Mao, Mingchao Hao, Ningbo Dai, Renzhe Xu, Shuyang Li, Tianyang Zhang, Yue He, Yuanrui Wang, Yunjia Zhang, Zijing Xu, Dongzhe Li, Fang Gao, Hao Zou, Jiandong Liu, Jiashuo Liu, Jiawei Xu, Kaijie Cheng, Kehan Li, Linjun Zhou, Qing Li, Shaohua Fan, Xiaoyu Lin, Xinyan Han, Xuanyue Li, Yan Lu, Yuan Xue, Yuanyuan Jiang, Zimu Wang, Zhenlei Wang, and Peng Cui.
\newblock Limix: Unleashing structured-data modeling capability for generalist intelligence.
\newblock \emph{CoRR}, abs/2509.03505, September 2025{\natexlab{a}}.

\bibitem[Zhang et~al.(2025{\natexlab{b}})Zhang, Maddix, Yin, Erickson, Ansari, Han, Zhang, Akoglu, Faloutsos, Mahoney, Hu, Rangwala, Karypis, and Wang]{zhang2025mitra}
Xiyuan Zhang, Danielle~C. Maddix, Junming Yin, Nick Erickson, Abdul~Fatir Ansari, Boran Han, Shuai Zhang, Leman Akoglu, Christos Faloutsos, Michael~W. Mahoney, Cuixiong Hu, Huzefa Rangwala, George Karypis, and Bernie Wang.
\newblock Mitra: Mixed synthetic priors for enhancing tabular foundation models.
\newblock In \emph{The Thirty-ninth Annual Conference on Neural Information Processing Systems}, 2025{\natexlab{b}}.

\bibitem[Zheng et~al.(2018)Zheng, Aragam, Ravikumar, and Xing]{NOTEARS}
Xun Zheng, Bryon Aragam, Pradeep Ravikumar, and Eric~P. Xing.
\newblock Dags with no tears: Continuous optimization for structure learning.
\newblock In \emph{Neural Information Processing Systems}, 2018.

\end{thebibliography}

\newpage

\appendix



\end{document}